\documentclass[11pt,a4paper]{article}
\usepackage[hyperref]{emnlp-ijcnlp-2019}
\usepackage{times}
\usepackage{latexsym}

\usepackage{url}
\usepackage{graphics}
\usepackage{graphicx}
\usepackage{multirow}
\usepackage{tabularx}
\usepackage{subcaption}

\usepackage{array}

\usepackage{amsmath}

\usepackage{amssymb}
\usepackage{pifont}
\newcommand{\cmark}{\ding{51}}%
\newcommand{\xmark}{\ding{55}}%

\aclfinalcopy 

\graphicspath{{./images/}}

\title{Exploring Deep Neural Networks and Transfer Learning for Analyzing Emotions in Tweets}

\author{Yasas Senarath \\
  University of Moratuwa, \\
  Sri Lanka \\
  \texttt{wayasas.13@cse.mrt.ac.lk} \\\And
  Uthayasanker Thayasivam \\
  University of Moratuwa, \\
  Sri Lanka \\
  \texttt{rtuthaya@cse.mrt.ac.lk} \\}

\date{}

\begin{document}
\maketitle
\begin{abstract}
In this paper, we present an experiment on using deep learning and transfer learning techniques for emotion analysis in tweets and suggest a method to interpret our deep learning models. The proposed approach for emotion analysis combines a Long Short Term Memory (LSTM) network with a Convolutional Neural Network (CNN). Then we extend this approach for emotion intensity prediction using transfer learning technique. Furthermore, we propose a technique to visualize the importance of each word in a tweet to get a better understanding of the model. Experimentally, we show in our analysis that the proposed models outperform the state-of-the-art in emotion classification while maintaining competitive results in predicting emotion intensity.
\end{abstract}

\section{Introduction}

Emotion analysis of user-generated content (UGC) available on the web provides insights toward making meaningful decisions. Micro-blog platforms such as Twitter has gained profuse popularity for textual content holding people's opinions. The past decade has seen the active growth in emotion analysis models in many domains. Recently there has been an increasing interest in analysis of emotions of informal short texts such as tweets. In this paper, we introduce and analyze a system to accurately identify the emotions of the individual tweets with the associated intensities~\footnote{Intensity refers to the degree or amount of an emotion}.

Analyzing emotions in social media such as twitter benefits society in a number of ways. Policymakers can use emotional information in social media to accurately identify concerns of people when making decisions. Monitoring social media for health issues benefits not only public health but also government decision makers~\cite{ji2013monitoring}. Furthermore, organizations can monitor opinion of the public on their products and services to provide better service to the society. Once emotions are recognized, emotion intensity can be used to prioritize the major concerns.

Studies in emotion analysis have often focused on emotion classification. However, emotions may exhibit varying levels of intensities. Here, emotion intensity can be defined as the degree or the intensity of particular emotion felt by the speaker. Additionally, we may observe multiple emotions simultaneously in the same tweet with varying intensities~\cite{bradley1992remembering}. 

One purpose of this study is to develop a model to accurately identify the emotions and associated emotion intensities for a given tweet. In this paper, we propose a transfer learning approach backed by a neural network classifier and a regressor. Although the proposed neural network alone is inadequate to beat the benchmark, we show that features learned when training the above neural networks can be used to improve the overall performance when combined with other features.

Another purpose of this study is to explain how the input word level features affect the features extracted by the neural network. 
The findings should make an important contribution in understanding how features are used in a neural network and to effectively select features to improve the effectiveness of extracted features. 

Our main contributions of this study:
\begin{itemize}
  \item Introduction of simpler but effective models for emotion classification and intensity prediction
  \item Apply state-of-the-art interpretation models to visualize and explain deep models for emotion intensity prediction
\end{itemize}

\pagebreak

Major challenge in using deep learning to train emotion intensity prediction models is the lack of large labeled datasets. More recently, emoji~\cite{felbo2017using} and hashtags~\cite{mohammad2015using} were used in studies to create large naturally labeled datasets. However, it is not possible to use a similar technique to obtain the intensity of emotions. Furthermore, creating a large dataset manually is time consuming and expensive. \cite{mohammad2017wassa, mohammad2018semeval} are some existing datasets for emotional intensity prediction. Due to the limited amount of task-specific training data the previous researches have opted for transfer learning approaches~\citet{baziotis2018ntua, duppada2018seernet} and traditional machine learning~\cite{kuijper2018ug18}. However, in this paper we argue that even with reasonable size dataset we can train a neural network to obtain good performance provided that there is proper regularization. Additionally, we show that features learned when training the neural network can be combined with other features to improve the overall performance of emotion intensity prediction. 


\begin{table}[]
\centering
\caption{The number of tweets in the SemEval-2018 Affect in Tweets Dataset}
\label{table:dataset-details}
\resizebox{0.4\textwidth}{!}{%
\begin{tabular}{lllll}
\hline
\textbf{Dataset} & \textbf{Train} & \textbf{Dev} & \textbf{Test} & \textbf{Total} \\ \hline
E-c & 6,838 & 886 & 3,259 & 10,983 \\
EI-reg &  &  &  &  \\
\hspace*{4pt}anger & 1,701 & 388 & 1,002 & 3,091 \\
\hspace*{4pt}fear & 2,252 & 389 & 986 & 3,627 \\
\hspace*{4pt}joy & 1,616 & 290 & 1,105 & 3,011 \\
\hspace*{4pt}sadness & 1,533 & 397 & 975 & 2,905 \\ \hline
\end{tabular}%
}
\end{table}

In \S~\ref{section:litreview}, we outline related works on sentiment and emotion mining. Next, in \S~\ref{section:data} we will discuss the datasets used in this study. After, we introduce the background and our methodology in \S~\ref{section:background} and \S~\ref{section:methodology} accordingly. Then, in \S~\ref{section:eval} we will discuss the evaluation results. Finally, we will conclude this paper in \S~\ref{section:conclusion}.

\section{Related Work}
\label{section:litreview}

Sentiment Analysis has become an important area, particularly when trying to analyze social media. Early examples of research into sentiment analysis involve in polarity classification of the textual input~\cite{pang2002thumbs, tang2014learning, dong2014adaptive, radford2017learning}. In recent years, there has been an increasing amount of literature on algorithms for emotion analysis which are closely aligned with our work. \citet{eisner2016emoji2vec} has introduced emoji2vec, a method to obtain emoji embedding from existing pre-trained word2vec~\cite{mikolov2013efficient} using emoji definitions. They have shown the importance of their emoji embedding by using it in sentiment analysis task. DeepMoji~\cite{felbo2017using} is a neural network trained on large twitter corpus naturally labeled for emoji. They have used a deep neural network with hidden bi-directional long short term memory layers (bi-LSTM) and an attention layer. This study has shown that transfer learning from DeepMoji can improve the performance of emotion and sentiment classification.

Several studies have investigated the approaches to predict the intensity of emotions in tweets. \citet{rosenthal2015semeval, kiritchenko2016semeval} have laid the groundwork for determining sentiment intensity of English phrases by introducing it as a shared task. Following the work of \citet{kiritchenko2016semeval}, \citet{mohammad2017emotion}~and~\citet{mohammad2018semeval} have introduced datasets for emotion intensity prediction in tweets. Various studies have been carried out on creating models for predicting emotion intensities of tweets. The model presented by \citet{duppada2018seernet} has achieved a 79.9\% Pearson correlation score in emotion intensity dataset presented in~\cite{mohammad2018semeval}. Their model composed of a stacked ensemble of xgboost regressors and random forest regressors via a meta-regressor and currently holds the benchmark results. In this approach, each base regressor is trained with a specific set of features transferred from a pre-trained model. This prohibits the ability to combine transfer features from multiple pre-trained models at the initial levels.
 
A number of published studies try to utilize Deep Neural Networks (DNNs) to analyze emotions\cite{baziotis2018ntua}. Up to now, far too little attention has been paid to explain those models. \citet{mohammad2018semeval} has studied how a number of systems perform under different biases. However, there are no insights on how input features are being used to make the prediction. \citet{baziotis2018ntua} has developed a methodology based on a deep attentive RNNs and transfer learning to analyze emotions while presenting the weight given by the self-attention mechanism as a viable solution to visualizing the word level importance. However, this does not provide holistic view of entire model.

\section{Dataset}
\label{section:data}
In this study, we utilize SemEval 2018 - Affect in Tweet task datasets to train and evaluate our models~\cite{mohammad2018semeval}. Specifically, we use datasets provided for emotion intensity regression task (EI-reg) and emotion classification task (E-c). 

\textbf{Emotion Classification}: Each tweet in this dataset is annotated for presence/absence of 11 emotions. List of annotated emotions: anger, anticipation, disgust, fear, joy, love, optimism, pessimism, sadness, surprise, trust. 

\textbf{Emotion Regression}:There are four separate sub-datasets for four emotions: anger, fear, joy and sadness. Each tweet is annotated with the intensity of the given emotion in the sub-dataset. 

Table~\ref{table:dataset-details} provides the number of training examples allocated for each partition in the dataset.

\section{Background}
\label{section:background}

\subsection{Long Short Term Memory (LSTM)}
Long Short Term Memory (LSTM)~\cite{hochreiter1997long} is an artificial network architecture developed to overcome the drawbacks of simple Recurrent Neural Networks (RNNs). Equation~\ref{eq:lstm-wiki} provides the formulation of the variant of LSTM that we are using for the experiments in this paper.

\begin{equation}
\label{eq:lstm-wiki}
\begin{split}
f_t &= \sigma_g (W_f x_t + U_fh_{t-1}+b_f) \\
i_t &= \sigma_g (W_i x_t + U_ih_{t-1}+b_i) \\
o_t &= \sigma_g (W_o x_t + U_oh_{t-1}+b_o) \\
c_t &= f_t \circ c_{t-1} + i_t \circ \sigma_c (W_c x_t + U_ch_{t-1}+b_c) \\
h_t &= o_t \circ \sigma_c (c_t) 
\end{split}
\end{equation}

Where variables $ x_t $, $ h_t $ represent input vector to the LSTM unit and hidden state vector accordingly. Weight matrices and bias vector parameters are indicated as $ W, U $ and $ b $. Activation functions indicated by $ \sigma_g $ , $ \sigma_c $ represents Sigmoid function and hyperbolic tangent function.

\subsection{Convolutional Neural Network (CNN)}
Convolutional Neural Networks~\cite{kim2014convolutional} can be used in a neural network to learn and extract important features from text. In this paper, we use pooling after each convolution operation to reduce the spatial size of the representation. 

\section{Methodology}
\label{section:methodology}
In this section, we will provide our approach and models for emotion analysis. First, we will discuss our base models for emotion intensity prediction with deep learning. Then we will describe our extended model based on xgboost regressor for emotion intensity prediction. Finally, we will describe the algorithms we used to explain the importance of word-level features on our base models.

\subsection{Preprocessing}
\label{section:preprocessing}
Tweets are processed before providing them as input to the classifier/ regressor.  We utilized \textit{ekphrasis}~\footnote{https://github.com/cbaziotis/ekphrasis}~\cite{baziotis2017datastories} for preprocessing with preprocessing steps: spell correction, word annotation, word segmentation and word tokenization.

\subsection{Base: Emotion Category Classification Unit (ECCU)}
\label{section:eccu}

\begin{figure}[t]
    \centering
    \includegraphics[scale=0.4]{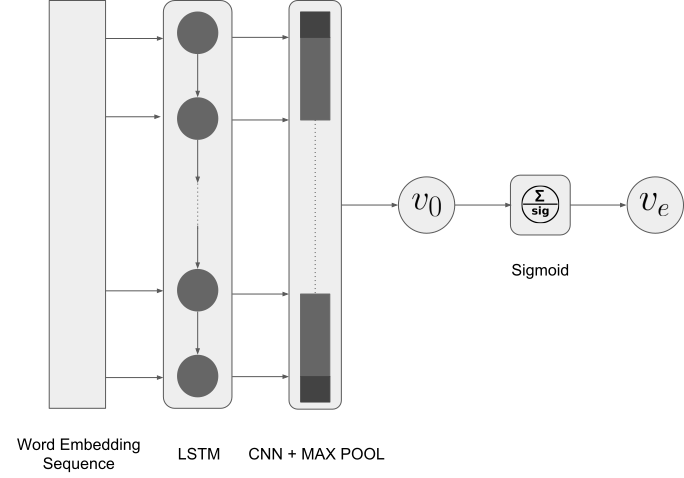}
    \caption{Emotion Category Classification Unit (ECCU) model for emotion classification.}
    \label{fig:rcnn_clf}
\end{figure}{}

\begin{figure}[t]
    \centering
    \includegraphics[scale=0.4]{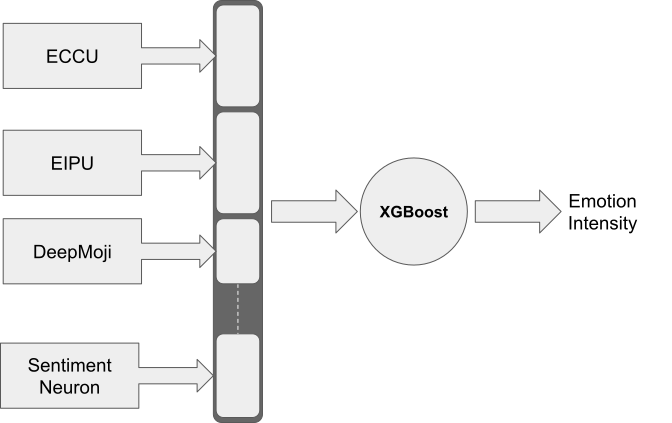}
    \caption{Emotion Intensity: Transfer Learning (EITL) model for intensity prediction.}
    \label{fig:xgboost_reg}
\end{figure}{}

Use of deep learning in text classification is becoming popular due to its robustness in automatic feature extraction. Here we use a Recurrent Convolutional Neural Network (RCNN) classifier with LSTM~\cite{hochreiter1997long} recurrent units. Figure~\ref{fig:rcnn_clf} illustrates the neural network architecture of ECCU.

Embedding sequence is fed to the LSTM layer using an embedding layer. At each time step (i.e. word) we obtain a vector from LSTM and fed it to a CNN layer. Then resulting vector is fed through a max pooling layer. The output vector of max-pool layer is identified as $v_{0}$. Finally, $v_{0}$ is passed through a Sigmoid layer to get the final emotion probability vector ($v_{e}$).

\subsection{Base: Emotion Intensity Prediction Unit (EIPU)}
\label{section:eipu}
Emotion Intensity Prediction Unit (EIPU) is similar in architecture to ECCU except it has only one output neuron with sigmoid activation. Therefore, there is only one predicted value which represents the emotional intensity for the emotion the network is trained for. 

\subsection{Emotion Intensity: Transfer Learning (EITL)}
\label{section:eitl}

\citet{pan2010survey} defines transfer learning as follows. Given a source domain $\mathcal{D}_S$ and learning task $\mathcal{T}_S$ and a target domain $\mathcal{D}_T$ and learning task $\mathcal{T}_T$, transfer learning aims to improve learning of the target predictive function $f_T(\cdot)$ in $\mathcal{D}_T$ using the knowledge in $\mathcal{D}_S$ and $\mathcal{T}_S$, where $\mathcal{D}_S \neq \mathcal{D}_T$ or $\mathcal{T}_S \neq \mathcal{T}_S$. 

In the EITL module, we use features obtained from multiple pretrained models with their own source domain and target to improve the performance of emotion intensity prediction task. Here we use XGBoost~\cite{chen2016xgboost} regressor as the target predictive function. As indicated in Figure 2, Features are combined to form a single vector by concatenating each transferred feature vector and then fed to the regressor. We experimented with features transferred from following pretrained models.

\begin{itemize}
  \item \textbf{ECCU features}: Union of features from the output of max pooling layer ($v_0$) and Sigmoid layer ($v_e$) of ECCU model. 
  \item \textbf{EIPU features}: Union of features from the output of max pooling layer ($v_0$) and Sigmoid layer ($v_e$) of EIPU model. 
  \item \textbf{DeepMoji features}: Union of features from attention layer and softmax layer of pre-trained DeepMoji model~\cite{felbo2017using}~\footnote{https://github.com/bfelbo/DeepMoji}.
  \item \textbf{Sentiment Neuron}: Features from pre-trained unsupervised sentiment neuron model~\footnote{https://github.com/openai/generating-reviews-discovering-sentiment}~\cite{radford2017learning}.
\end{itemize}

\begin{table}[]
\centering
\caption[Model and training hyper-parameters for ECCU and EIPU models]{Model and training hyper-parameters for ECCU and EIPU models. *Number of epochs for Anger emotion intensity model is 40 while for all other emotions we used 15.}
\label{table:model-parameters-eccu}
\resizebox{0.5\textwidth}{!}{%
\begin{tabular}{llc}
\hline
\multicolumn{1}{c}{Parameter} & \multicolumn{1}{c}{ECCU} & EIPU \\ \hline
LSTM &  & \multicolumn{1}{l}{} \\
\hspace*{8pt}Units & 128 & 64 \\
\hspace*{8pt}Dropout Rate & 0.5 & 0.8 \\
Convolutional Layer &  & \multicolumn{1}{l}{} \\
\hspace*{8pt}Filters & 128 & 64 \\
\hspace*{8pt}Kernel size & 2 & 2 \\
\hspace*{8pt}Padding & Same & Same \\
\hspace*{8pt}Activation & ReLU & ReLU \\
Dropout Layer &  & \multicolumn{1}{l}{} \\
\hspace*{8pt}Rate & 0.5 & 0.8 \\
Last (Dense) Layer &  & \multicolumn{1}{l}{} \\
\hspace*{8pt}Activation & Sigmoid & Sigmoid \\
Training &  & \multicolumn{1}{l}{} \\
\hspace*{8pt}Number of Epochs & 10 & 15/40* \\
\hspace*{8pt}Batch Size & 8 & 8 \\ \hline
\end{tabular}
}
\end{table}

\begin{table}[ht]
\centering
\caption[Parameters used for training EITL model]{Parameters used for training EITL model. C1 indicates parameters for Anger, Joy, Sadness emotion intensity models. C2 indicates training parameters for Fear emotion intensity prediction model.}
\label{table:model-parameters-eitl}
\begin{tabular}{lcc}
\hline
\multicolumn{1}{c}{\textbf{Parameter}} & \textbf{C1} & \textbf{C2} \\ 
\hline
Features & \multicolumn{1}{l}{} & \multicolumn{1}{l}{} \\
\hspace*{8pt}DeepMoji features & \cmark & \cmark \\
\hspace*{8pt}Sentiment Neuron & \cmark & \cmark \\
\hspace*{8pt}ECCU features & \cmark & \cmark \\
\hspace*{8pt}EIPU features & \cmark & \xmark \\
Max Depth & 2 & 5 \\
Learning Rate & 0.01 & 0.01 \\
\# of Estimators & 400 & 300 \\
\hline
\end{tabular}
\end{table}

\begin{table*}[ht]
\centering
\caption{Performance scores of emotion intensity prediction models. The marker $\dagger$ indicates the benchmark~\cite{duppada2018seernet} and * indicates results obtained in~\cite{baziotis2018ntua}}
\label{table:reg_score}
\begin{tabular}{lrrrrr}
\hline
\multicolumn{1}{c}{\multirow{2}{*}{\textbf{Model}}} & \multicolumn{4}{c}{\textbf{Pearson Correlation}} & \multirow{2}{*}{\textbf{Average}} \\ \cline{2-5}
\multicolumn{1}{c}{} & \multicolumn{1}{c}{Anger} & \multicolumn{1}{c}{Fear} & \multicolumn{1}{c}{Joy} & \multicolumn{1}{c}{Sadness} &  \\ \hline
EIPU & 76.45\% & 67.08\% & 72.10\% & 68.95\% & 70.83\% \\ 
EITL & 82.16\% & \textbf{78.67\%} & 78.42\% & \textbf{79.99\%} & 79.81\% \\
SeerNet$^\dagger$ & \textbf{82.70\%} & 77.90\% & \textbf{79.20}\% & 79.80\% & \textbf{79.90\%} \\ 
NTUA-SLP$^*$ & 78.20\% & 75.80\% & 77.10\% & 79.80\% & 77.70\% \\ 
\hline
BoW & 52.49\% & 52.27\% & 57.16\% & 47.21\% & 52.28\% \\ 
NBoW & 65.39\% & 63.18\% & 63.55\% & 63.05\% & 63.79\% \\ 
NBoW+A & 65.60\% & 63.59\% & 63.84\% & 63.41\% & 64.11\% \\ \hline
\end{tabular}%
\end{table*}

\begin{table}[]
\caption{Performance scores for ECCU compared with the benchmark systems. The marker $\dagger$ indicates the benchmark~\cite{baziotis2018ntua}.}
\label{table:clf_score}
\resizebox{0.5\textwidth}{!}{%
\begin{tabular}{lrrrr}
\hline
\multicolumn{1}{c}{\textbf{Model}} & \multicolumn{1}{c}{\textbf{\begin{tabular}[c]{@{}c@{}}Accuracy\\ (jaccard)\end{tabular}}} & \multicolumn{1}{c}{\textbf{\begin{tabular}[c]{@{}c@{}}F1 \\ (Micro)\end{tabular}}} & \multicolumn{1}{c}{\textbf{\begin{tabular}[c]{@{}c@{}}F1 \\ (Macro)\end{tabular}}} \\ \hline
ECCU & \textbf{58.63} & \textbf{71.92} & \textbf{52.8} \\ 
NTUA-SLP$^\dagger$ & 57.88 & 70.1 & \textbf{52.8} \\ 
Rnd. Baseline & 18.5 & 30.7 & 28.5 \\ \hline
\end{tabular}
}
\end{table}

\begin{table*}[]
\caption{Examples for word level importance heat-map visualizations. Columns represented by letters E, A and P represents Emotion, true emotion intensity and predicted emotional intensity respectively. Letters A, F, J and S in column E corresponds to emotions Anger, Fear, Joy and Sadness. The predicted emotion intensity of tweets above the double-line separator is closer to the actual value while the difference between the actual and predicted is significantly higher for the tweets below that separator. }
\label{table:deep_explain}
\centering
\resizebox{\textwidth}{!}{%
\begin{tabular}{p{1cm}lll}
\hline
\\[-0.5em]
\huge{E} & \huge{Tweet} & \huge{A} & \huge{P} \\ \hline
\\
\huge{A} & \raisebox{-.5\height}{\includegraphics[scale=1]{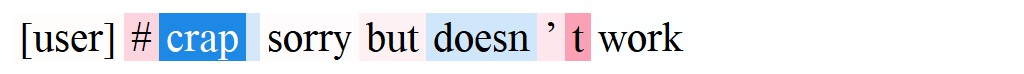}} & \huge{0.563} & \huge{0.556}  \\
\huge{A} & \raisebox{-.5\height}{\includegraphics[scale=1]{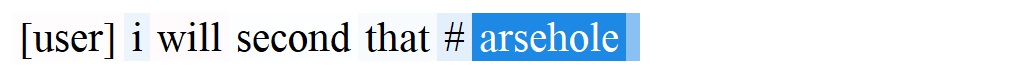}} & \huge{0.697} & \huge{0.690}  \\ \hline
\\
\huge{F} & \raisebox{-.5\height}{\includegraphics[scale=1]{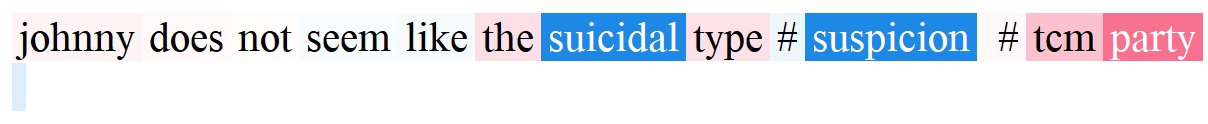}} & \huge{0.620} & \huge{0.610}  \\
\huge{F} & \raisebox{-.5\height}{\includegraphics[scale=1]{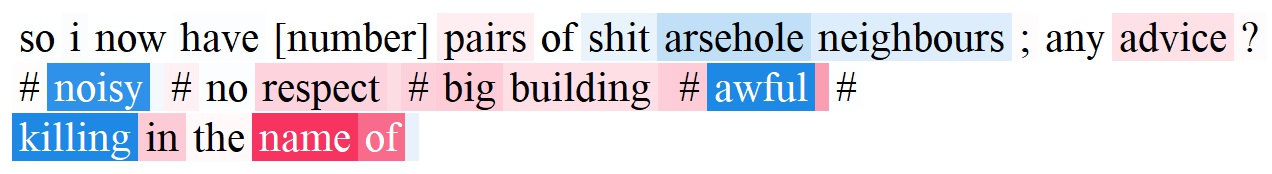}} & \huge{0.625} & \huge{0.576}  \\ \hline
\\
\huge{J} & \raisebox{-.5\height}{\includegraphics[scale=1]{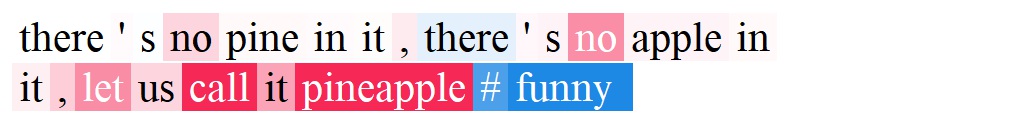}} & \huge{0.547} & \huge{0.537}   \\
\huge{J} & \raisebox{-.5\height}{\includegraphics[scale=1]{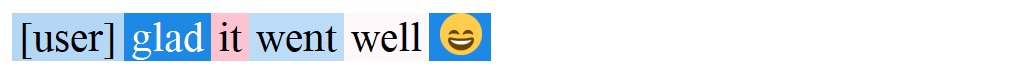}} & \huge{0.656} & \huge{0.647}   \\ \hline
\\
\huge{S} & \raisebox{-.5\height}{\includegraphics[scale=1]{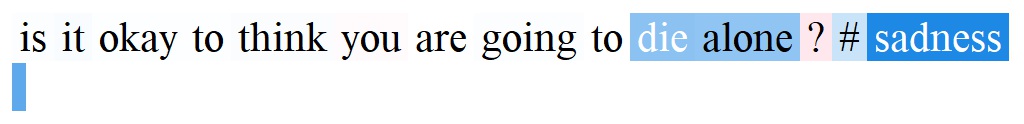}} & \huge{0.731} & \huge{0.723}   \\
\huge{S} & \raisebox{-.5\height}{\includegraphics[scale=1]{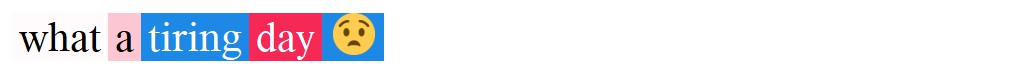}} & \huge{0.636} & \huge{0.6277}  \\ \hline \\[-0.8em] \hline
\\
\huge{A} & \raisebox{-.5\height}{\includegraphics[scale=1]{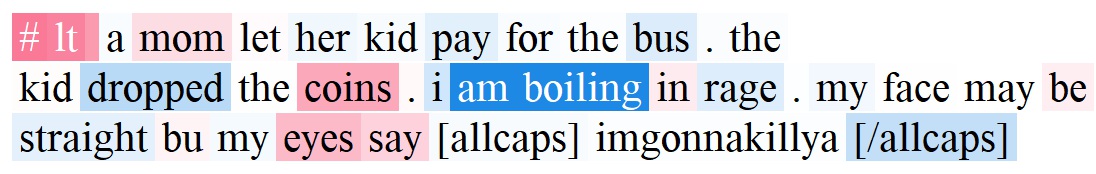}} &  \huge{0.848} & \huge{0.505}  \\ 
\huge{A} & \raisebox{-.5\height}{\includegraphics[scale=1]{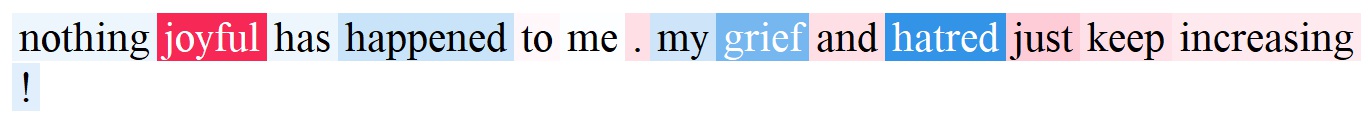}} &  \huge{0.813} & \huge{0.425}  \\ \hline
\\  
\huge{F} & \raisebox{-.5\height}{\includegraphics[scale=1]{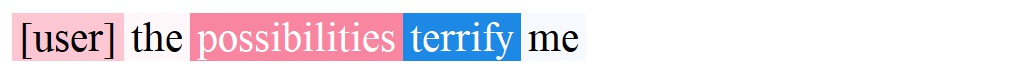}} &  \huge{0.911} & \huge{0.562}  \\ 
\huge{F} & \raisebox{-.5\height}{\includegraphics[scale=1]{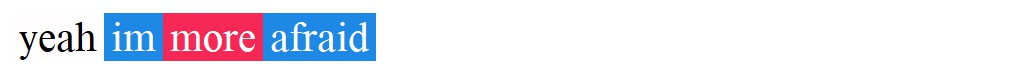}} &  \huge{0.913} & \huge{0.578}  \\ \hline
\\
\huge{J} & \raisebox{-.5\height}{\includegraphics[scale=1]{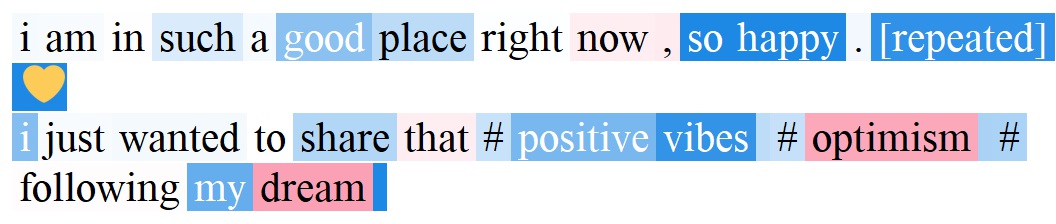}} &  \huge{0.955} & \huge{0.668}  \\ 
\huge{J} & \raisebox{-.5\height}{\includegraphics[scale=1]{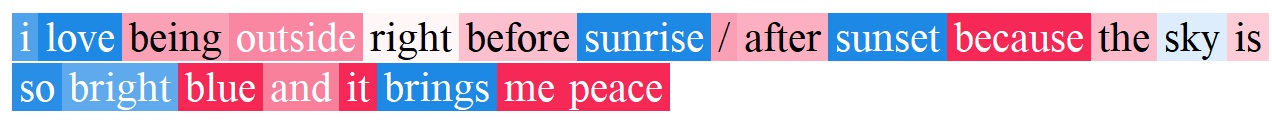}} &  \huge{0.828} & \huge{0.560}  \\ \hline
\\
\huge{S} & \raisebox{-.5\height}{\includegraphics[scale=1]{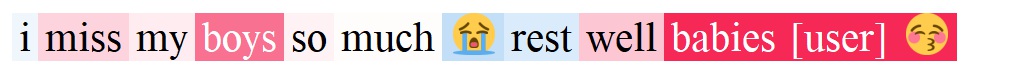}} &  \huge{0.696} & \huge{0.360}  \\ 
\\
\huge{S} & \raisebox{-.5\height}{\includegraphics[scale=1]{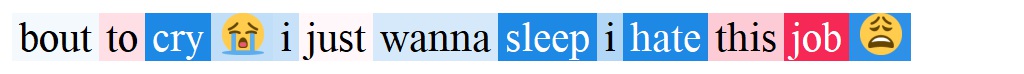}} &  \huge{0.786} & \huge{0.454}  \\ \hline
\end{tabular}%
}
\end{table*}

\subsection{Training Process}
The embedding layer is initialized with Twitter specific word2vec published in \cite{baziotis2018ntua}. We maintain the original word embedding by preventing the training algorithm from fine-tuning the word embedding layer. We used categorical cross-entropy loss and mean squared error as loss function when training ECCU and EIPU respectively. Adam optimizer is used since it can generate better results fast. The hyper-parameters for training the proposed network was based on results on validation dataset provided in SemEval Task~\cite{mohammad2018semeval}. 

Table~\ref{table:model-parameters-eccu} indicates the hyper-parameters and training parameters used for generating the models. We applied a dropout~\cite{srivastava2014dropout} layer after max-pooling layer in the proposed neural network to avoid overfitting while training. We have similar hyper-parameters and training parameters for each emotion in EIPU models. Table~\ref{table:model-parameters-eitl} shows the training parameters and the features used in the EITL model.

\subsection{Deep Explanations}
We experiment with using Deep SHAP (SHapley Additive exPlanations)~\cite{lundberg2017unified} to explain and visualize the importance of each word in a given tweet for the prediction of emotion intensity using EIPU model. The first step in this process was to obtain SHAP values for our end-to-end neural network using Deep SHAP. Then we obtain the normalized SHAP value ($I_i$) as the importance of $i^{th}$ word in a Tweet using Equation~\ref{eq:normaize}. $S$ is the SHAP vector for a given input (Tweet) and ${S_i}$ is the SHAP value of $i^{th}$ word.

\begin{equation}
\label{eq:normaize}
I_{i} = \frac{S_{i}}{max(abs(S))}
\end{equation}

\section{Evaluation}
\label{section:eval}

Table~\ref{table:clf_score} indicates the performance score for classification model using accuracy, micro F1 and macro F1 scores. We clearly see that ECCU outperforms the benchmark system and a random baseline. Here we obtained random baseline results by a system that randomly guesses the prediction.

In Table~\ref{table:reg_score} we compare our models against existing systems and three strong baselines. We obtained the baseline results from NTUA-SLP~\cite{baziotis2018ntua}. The first baseline is the unigram Bag-of-Word (BoW) model with TF-IDF weighting. Second baseline is the Neural BoW (NBoW) model, constructed by averaging the word2vec embedding of words in a Tweet. Last baseline is similar to NBoW except it has extra 10-dimensions in the embedding with affective information (NBoW+A). Aforementioned features are used as inputs to an SVM with C=0.6 to obtain the baselines.

We observe that our neural model (EIPU) outperforms the baselines with substantial performance improvement. Moreover, we see that our proposed transfer learning model outperforms the existing state of the art models for two emotions: fear and sadness while maintaining competitive results over other emotions. Additionally, we clearly exceed the NTUA-SLP~\cite{baziotis2018ntua}, the second best system at SemEval 2018 EI-reg subtask of Emotion in Tweet task. However, EIPU did not perform well with respect to the transfer learning based models in Table~\ref{table:reg_score}. This behaviour can be attributed to the extra information provided through transfer learning. 

Table~\ref{table:deep_explain} visualizes the word level importance when predicting the emotion intensity for some selected Tweets. The tweets are organized in sections based on emotion and  The blue color corresponds to a positive impact on the prediction while red color corresponds to words that help in decreasing the predicted emotion intensity. The intensity of color indicates the importance of that word in the final outcome. An important observation from this visualization is that certain keywords and emoji plays a major role in the final outcome while taking a good consideration on the context of the text.

\section{Conclusion}
\label{section:conclusion}
In this study, we propose a simple yet effective model for emotion classification and emotion intensity prediction in Tweets while suggesting a method to explain and visualize a trained DNN. We utilized a neural network with LSTM layer followed by a convolution layer with max-pooling for emotion category classification as well as emotion intensity prediction. We extend this work by transferring features from above models and two state-of-the-art models trained for different tasks to a XGBoost regressor to predict the emotion intensity in Tweets more accurately. Moreover, we suggest a technique to visualize and interpret the feature importance of trained DNNs for emotion intensity prediction. In the future, we plan on experimenting with using attentive mechanisms~\cite{vaswani2017attention} to improve the emotion intensity prediction further. Our models outperformed existing state-of-the-art models for emotion classification and in predicting fear and anger emotion intensities, while maintaining a competitive results in predicting other emotions.

\bibliographystyle{acl_natbib}
\bibliography{emnlp-ijcnlp-2019}

\end{document}